\documentclass{article}

\usepackage{arxiv}
\usepackage{tabularx}
\usepackage[utf8]{inputenc} 
\usepackage[T1]{fontenc}    
\usepackage{hyperref}       
\usepackage{url}            
\usepackage{booktabs}       
\usepackage{amsfonts}       
\usepackage{nicefrac}       
\usepackage{microtype}      
\usepackage{lipsum}		
\usepackage{graphicx}
\usepackage{natbib}
\usepackage{doi}
\usepackage{makecell}
\usepackage{comment}
\usepackage{amsmath}
\usepackage{algpseudocode}
\usepackage{amssymb}
\usepackage{makecell}
\usepackage{algorithm}

\title{QuaSAR: Quantization Compensation via Stable Activation-Aware Rank Truncation}

\author{ \href{https://orcid.org/0000-0000-0000-0000}{\includegraphics[scale=0.06]{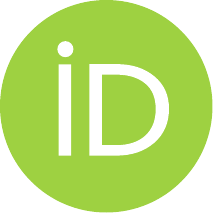}\hspace{1mm}Lin-Fa Lee}\\
	Department of Institute of Artificial Intelligence Innovation\\
	National Yang Ming Chiao Tung University\\
	Hsinchu, Taiwan \\
	\texttt{prologue.ii14@nycu.edu.tw} \\
	\And
	\href{https://orcid.org/0000-0000-0000-0000}{\includegraphics[scale=0.06]{orcid.pdf}\hspace{1mm}Yi-Yu Chang} \\
	Department of Institute of Artificial Intelligence Innovation\\
	National Yang Ming Chiao Tung University\\
	Hsinchu, Taiwan \\
	\texttt{daniel282907@gmail.com} \\
    \And
    \href{https://orcid.org/0000-0000-0000-0000}{\includegraphics[scale=0.06]{orcid.pdf}\hspace{1mm}Kuo-Hui Yeh} \\
	Department of Institute of Artificial Intelligence Innovation\\
	National Yang Ming Chiao Tung University\\
	Hsinchu, Taiwan \\
	\texttt{khyeh@nycu.edu.tw} \\
}

\renewcommand{\shorttitle}{WebMCP Tool Surface Poisoning}

\hypersetup{
pdftitle={A template for the arxiv style},
pdfsubject={q-bio.NC, q-bio.QM},
pdfauthor={David S.~Hippocampus, Elias D.~Striatum},
pdfkeywords={First keyword, Second keyword, More},
}

\begin{document}
\maketitle

	Recent training-free post-training quantization methods restore model accuracy through closed-form residual compensation. To constrain additional model storage overhead, several existing methods gate layer selection by goodness-of-fit, retaining only those layers whose compensation yields a positive residual fit score and discarding the rest. In this paper, we show that, under the low-bit W4A4 setting, this gating mechanism fails to distinguish poorly predictable quantization error from numerical solver failure. Rank-deficient input activations yield severely ill-conditioned or numerically singular Gram matrices, causing the closed-form solver to become unstable and produce spuriously negative fit scores. Consequently, existing goodness-of-fit gates misclassify affected layers as uncompensable and discard them. Many of these discarded layers can nevertheless provide substantial error recovery when their compensation is computed using a numerically stable solver. To address this problem, we propose a parameter-free truncated pseudoinverse solver which removes collapsed directions prior to inversion. On ViT-B with the W4A4 setting, our training-free method achieves 81.42\% top-1 accuracy, outperforming prior post-training methods and fine-tuning-based baselines. Combined with joint low-rank and quantization compression, the proposed method reaches a deployable operating point of 80.26\% accuracy at 54.7 MB, providing a well-balanced trade-off between model size and accuracy.

\section{Introduction}

Quantization is a mainstream method for compressing models and facilitates deployment on resource constrained devices. Existing quantization methods are generally divided into two categories: quantization aware training (QAT) \cite{liu2024llm,chen2025efficientqat,zhao2023post} and post training quantization (PTQ) \cite{li2021brecq,arai2026quantization,bai2022towards}. The former requires a full dataset and additional model training, while the latter requires only a small calibration set. Therefore, PTQ is often more suitable for practical deployment scenarios in which training data or computational resources are limited.


In the training-free PTQ setting, recent studies model the quantization error as a linear function of the input, derive a compensation term in closed form, and add the resulting correction to the quantized output. Such methods can recover the model accuracy to a level close to that of the original full-precision model \cite{zhang2026quantvla,shang2023post,tang2025qwt}. This type of closed form compensation relies on the inversion of the activation covariance matrix, which is often ill conditioned under low bit quantization. 

Several techniques have been adopted to improve numerical stability. GPTQ applies dampening to the diagonal of the Hessian \cite{frantar2022gptq}, and some methods introduce ridge regularization \cite{hastie2020ridge}. However, these approaches mostly treat ill conditioning as an implementation detail and preemptively circumvent it, without examining the core issue of how ill conditioning leads to the actual failure of compensation, or even the erroneous discarding of originally compensable positions. The consequence is that the value of compensation is not realized in the positions where it is needed most, while the compensation module itself increases the model size. The failure is concrete and measurable. A collapsed layer returns a mathematically impossible $R^2 < 0$ on its own fitted data, which the common $R^2 > 0$ gate reads as a verdict of low value rather than as an alarm of numerical collapse. On ViT-B/W4A4 the layer this gate most confidently discards is $b0.\mathrm{fc2}$, reporting $R^2 = -68.6$ and a compensation parameter norm of $6.2\times10^{7}$. It is also the single most valuable compensator in the model: removing it costs $1.05\%$ top 1, more than any other layer. The gate discards precisely what it should keep.

Notably, existing studies on closed-form low-rank compensation have mainly considered weight-only quantization for large language models, such as QERA, CALDERA, and GPTQ-iLoRA. However, efficient inference for vision models often requires both weights and activations to be quantized, particularly under the W4A4 setting, as considered in QServe and MixA-Q. Compared with weight-only quantization, W4A4 is substantially more sensitive to accuracy degradation. Moreover, the high dimensional activations of certain layers are intrinsically redundant, rendering the closed form compensation regression fragile by nature. Activation quantization acts as an amplifier that pushes a subset of layers past the threshold of numerical collapse, which is an issue not encountered in weight only compensation research.

Specifically, existing residual compensation faces an unresolved trade off under this setting. QwT \cite{fu2025quantization} trades full compensation for accuracy at the cost of model size, while its successor QwT-v2 \cite{tang2025qwt} utilizes static diagonal compensation to achieve an extremely lightweight size but sacrifices compensation capability. There lacks an operating point that balances both accuracy and size between the two. This paper proposes a training free layer wise compensation framework and makes three contributions:

\begin{itemize}
\item \textbf{A field wide diagnosis, not a new estimator.}
We identify an overlooked failure mechanism in W4A4 compensation. The
closed form solution is intrinsically fragile on rank deficient
activation structures, and activation quantization amplifies this
fragility into numerical collapse, leading such layers to be misjudged
as uncompensable and thereby discarding originally high value
compensation layers. We further show the trigger is rank deficiency
rather than ill conditioning per se. This failure occurs across both
CNN- and Transformer-based architectures.

\item \textbf{Recovering the layers the gate discards.}
We introduce a parameter-free truncated pseudoinverse that removes only
the collapsed directions while preserving the remaining compensable
subspace. By recovering the erroneously discarded high value layers through a
parameter free truncated pseudoinverse that removes only the collapsed
directions, our method achieves $81.42\%$ on ViT-B/W4A4 without
additional training, surpassing all prior training free methods and even the
fine tuning based QwT$^{\ast}$.

\item \textbf{A deployable operating point, and how to budget it.}
Through low rank decomposition and compensation quantization, we
further compress the compensation to a deployable scale, achieving an
operating point of $80.26\%$ accuracy at $54.7\,\mathrm{MB}$. This
outperforms the fully compensated QwT and the extremely lightweight
QwT-v2, filling the missing balance between accuracy and size. We
further establish that the compensation budget should be allocated
based on the global scale rather than in a fine grained, position wise
manner. Reversing a high rank to important layers allocation to its
exact opposite changes accuracy by only $0.02\%$, showing that layer
importance does not predict rank sensitivity.

\end{itemize}
\section{Related Work}

\subsection{PTQ for Vision Transformers}
Post training quantization (PTQ) deploys resource intensive networks on
edge devices without the retraining pipeline of quantization aware
training (QAT) \cite{mohammadi2026fixing}. Early PTQ methods achieved
high fidelity on convolutional networks, but their direct application to
Vision Transformers (ViTs) causes severe degradation
\cite{moon2024instance}, driven by the extreme inter channel variance in
post LayerNorm activations and the heavy tailed distributions of post
Softmax attention maps \cite{li2023repq}. Targeted frameworks address
these distributions from two directions. RepQ-ViT decouples the
calibration time quantization grid from the hardware friendly quantizer
executed at inference \cite{li2023repq}, and IGQ-ViT partitions
activation channels into instance aware groups to isolate token dependent
outliers \cite{moon2024instance}. Reconstruction based methods instead
refine quantization parameters through localized optimization, using
Fisher information approximations \cite{wu2025fima} or average
perturbation Hessians \cite{wu2025aphq}. All remain constrained by the
representational limits of uniform low bit formats, leaving substantial
quantization noise uncorrected under aggressive settings such as W4A4.

\subsection{Low Rank Error Reconstruction}
A parallel line of work augments low precision backbones with high
precision low rank components, exploiting the low rank structure of
quantization error matrices \cite{zhang2025qera}. For parameter efficient
fine tuning, LoftQ jointly initializes the quantized weights and the low
rank adapters to approximate the pre trained full precision weights
\cite{li2024loftq}. For post training scenarios, QERA derives a closed
form solution minimizing output activation discrepancy via singular value
perturbation theory \cite{zhang2025qera}, CALDERA compresses both the
backbone and the low rank matrices by alternating minimization
\cite{saha2024compressing}, and SVDQuant smooths activation outliers into
the weight matrices and absorbs them into a 16 bit low rank branch
\cite{zhang2026q}. These analytical methods are biased toward weight only
quantization or rely on parallel full precision branches that incur
runtime and memory bottlenecks. Crucially, all of them assume stable,
high precision activation inputs and do not examine the mathematical
vulnerability of the underlying solver when the activation matrices are
themselves heavily quantized and rank deficient.

\subsection{Structural Residual Compensation}
To avoid the overhead of parallel high precision branches, recent
compensation paradigms apply lightweight structural modifications that
directly correct activation discrepancies \cite{mohammadi2026fixing}.
Quantization without Tears (QwT) appends lightweight linear layers in
parallel with quantized blocks and solves a closed form least squares
regression on a calibration set to predict and add back the residual
quantization error \cite{fu2025quantization}. It recovers substantial
accuracy without backpropagation but introduces a parameter footprint of
roughly 30\% and requires parallel floating point computation
incompatible with integer only hardware \cite{tang2025qwt}. QwT-v2
replaces the full linear projections with channel wise affine
compensation that folds into the existing quantization scales and zero
points, achieving zero deployment overhead \cite{tang2025qwt}. This
simplification to diagonal operators limits representational capacity and
leaves a noticeable gap under W4A4. Both methods treat the inversion of
the activation Gram matrix as an implementation detail, applying diagonal
dampening or ridge regularization without analyzing the physical cause of
the instability.
\section{Proposed Method}
\begin{figure}[!t]
    \centering
    \includegraphics[
        width=1\columnwidth,
        height=1\textheight,
        keepaspectratio
    ]{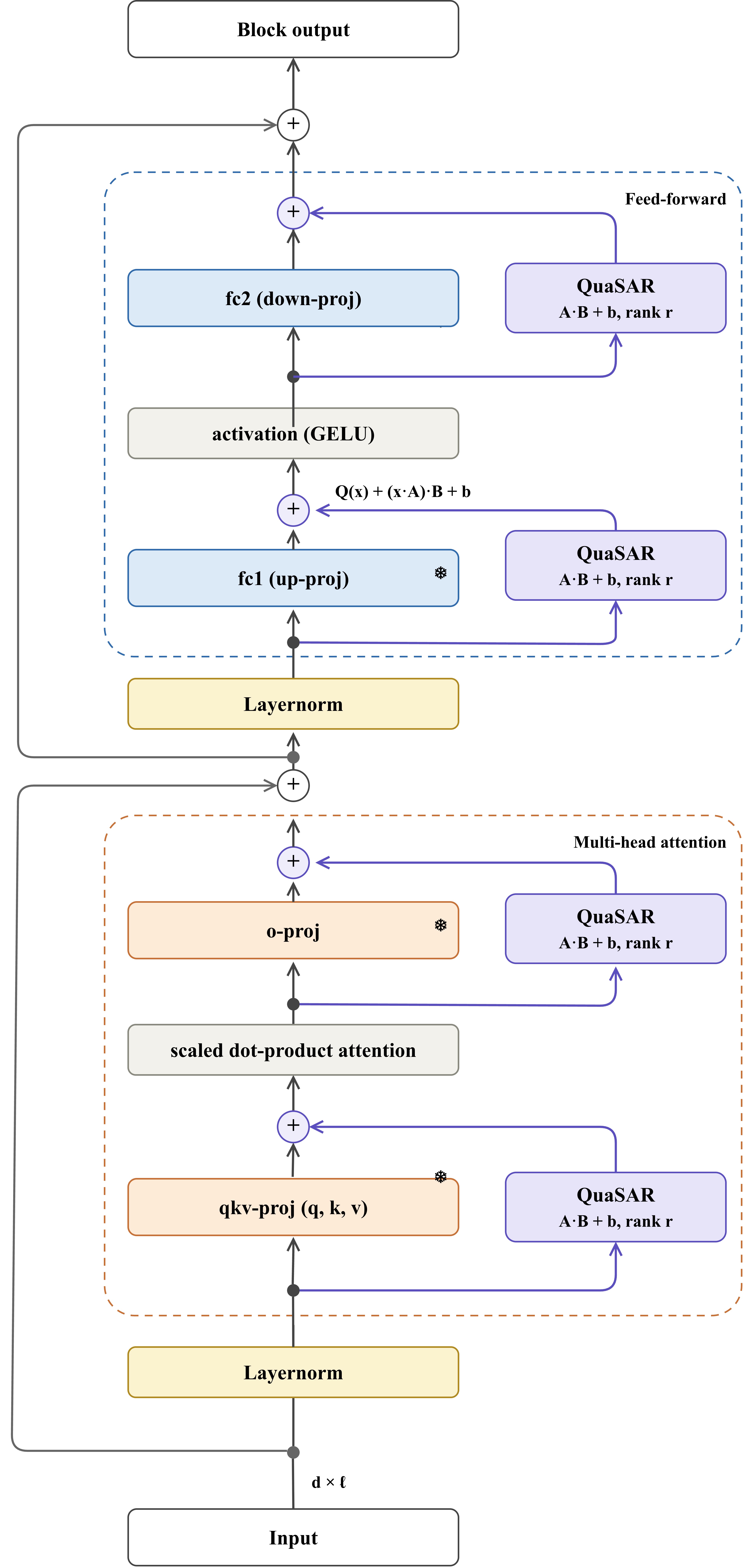}
    \caption{}
    \label{fig:qwtv3a}
\end{figure}

\begin{figure*}[!t]
    \centering
    \includegraphics[width=\textwidth]{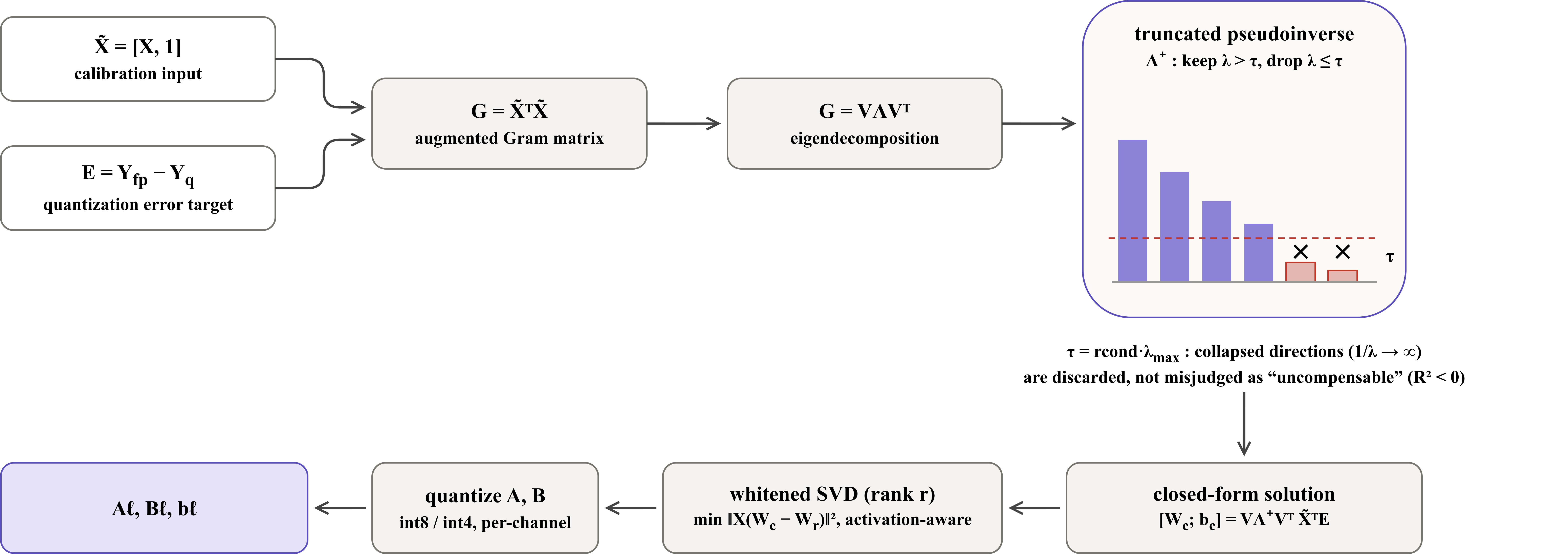}
    \caption{}
    \label{fig:qwtv3b}
\end{figure*}

We consider the post training quantization (PTQ) of Transformers under the W4A4 setting (where both weights and activations are quantized to 4 bit). For a given linear layer, let its full precision output be $Y_{\text{fp}}$ and its actual output after W4A4 quantization be $Y_{\text{q}}$. The quantization error is the difference between the two:
\begin{equation}
    E = Y_{\text{fp}} - Y_{\text{q}}
\end{equation}

The core idea of residual compensation is to linearly predict this error using the input $X$ of the layer, and to add the prediction back to the quantized output. If the compensation parameters $(W_{\text{c}}, b_{\text{c}})$ are obtained such that $X W_{\text{c}} + b_{\text{c}} \approx E$, the deployed output of the layer becomes:
\begin{equation}
    Y_{\text{deploy}} = Y_{\text{q}} + (X W_{\text{c}} + b_{\text{c}}) \approx Y_{\text{fp}}
\end{equation}

We formulate this as a least squares problem. The complete framework, as illustrated in Figure \ref{qwt v3-Panel (a) block (1)_0} and Figure \ref{qwt v3-Panel (b) solver (1)_0} and Algorithms~1 and~2 (at Appendix), consists of three components: layer wise compensation, numerically stable solving, and dual axis compression, detailed in the following subsections respectively.

\subsection{Layer wise Compensation}

As shown in Figure \ref{qwt v3-Panel (a) block (1)_0} and Figure \ref{qwt v3-Panel (b) solver (1)_0}, we push down the compensation to the individual linear layers within the Transformer block, attaching an independent compensator to each of the four linear layers, rather than sharing a single compensator per block. Independent compensation allows the error at each layer to be corrected more precisely, contributing an accuracy gain of $+3.1\%$ over block wise compensation in our experiments. Each compensator is deployed in parallel with its corresponding linear layer, formulated as:
\begin{equation}
 \text{output} = Q(x) + (x \cdot A) \cdot B + b 
\end{equation}
where $A$ and $B$ are low rank factors detailed in the last subsection.

Compensating a specific layer alters its output, which in turn affects the actual input to downstream layers. If we were to capture the pre compensation inputs for all layers at once to solve for the parameters, it would cause a mismatch with the deployment phase. Therefore, we sequentially compensate each layer (Algorithm~2 at Appendix), immediately replace it with the compensated version and plug it back into the model, and then capture the true input for the next layer to solve. Concretely, for each layer $\ell$ we feed the same actual input $X$, which includes all upstream quantization and compensation effects, through both the full precision weights and the quantized weights of $\ell$; the error target $E$ is the difference between these two outputs, so each compensator corrects the local error of its own layer under the true deployment time input distribution.

\subsection{Numerically Stable Compensation Solution}

We point out that the aforementioned closed form solution suffers from a numerical failure during compensation under the W4A4 setting, an issue overlooked by existing methods, and propose a stable solver (Algorithm~1, lines 2--5).

The least squares objective for compensation is:
\begin{equation}
 \min \| X W_{\text{c}} + b_{\text{c}} - E \|^2 
\end{equation}
By absorbing the bias into the augmented input $\tilde{X} = [X, 1]$, the solution is given by:
\begin{equation}
 \begin{bmatrix} W_{\text{c}} \\ b_{\text{c}} \end{bmatrix} = G^{-1} \tilde{X}^T E \quad \text{where} \quad G = \tilde{X}^T \tilde{X} 
\end{equation}
(corresponding to Algorithm~1, line 2). We find that the numerical issues are concentrated in the inversion of $G$.

Given that $G$ is symmetric positive semi definite, its eigendecomposition is $G = V \Lambda V^T$, and its inverse is $G^{-1} = \sum_i \frac{1}{\lambda_i} v_i v_i^T$, i.e., taking the reciprocal of each eigenvalue. The activations of certain layers are intrinsically redundant: their effective dimensionality is far below the nominal width, so $G$ is inherently near rank deficient even under full precision inputs. Under the W4A4 setting, activation quantization further compresses slightly different input dimensions into nearly identical ones, reducing the effective rank and pushing such layers past the threshold of collapse. During inversion, these near zero directions produce reciprocals $1/\lambda_i$ approaching infinity, causing the solved compensation parameters to explode and the entire compensation to collapse.

The direct fingerprint of collapse is rank deficiency: the Gram matrix of a collapsed layer becomes numerically singular, with its smallest eigenvalue indistinguishable from zero at float32 precision, falling within the eigenvalue noise floor $\approx \epsilon \cdot \lambda_{\max}$. This rank deficiency manifests through two robust signals: the compensation parameter norm $\|W_c\|$ explodes by several orders of magnitude, and the goodness of fit $R^2$ turns sharply negative (Appendix~A.1). As a coarse gauge of ill conditioning we also report the condition number:
\begin{equation}
\kappa(G) = \frac{\lambda_{\max}}{\lambda_{\min}^{+}}
\end{equation}
where $\lambda_{\min}^{+}$ denotes the smallest eigenvalue above the numerical noise floor; eigenvalues within the float32 noise floor are excluded, as their magnitudes are not meaningful.

Empirical measurements reveal that $\kappa(G)$ for ViT-B can reach $10^7$--$10^9$; a large $\kappa$ alone, however, does not predict collapse. The layer with the largest $\kappa$ in the entire model does not collapse, whereas every collapsed layer exhibits a numerically singular Gram matrix, identified by an exploding $\|W_c\|$ and a sharply negative $R^2$. According to existing methods, the criterion $R^2 > 0$ used to determine ``whether the layer is worth compensating'' treats this as an uncompensable numerical error. Here:
\begin{equation}
 R^2 = 1 - \frac{\| E - (X W_{\text{c}} + b_{\text{c}}) \|^2}{\| E - \bar{E} \|^2} 
\end{equation}
Under normal solving conditions, the least squares solution must have $R^2 \ge 0$ on its fitted data. However, the meaningless parameters solved during a numerical collapse yield mathematically impossible $R^2 < 0$ on the same batch of data, leading them to be misjudged as ``uncompensable'' by this threshold and subsequently discarded. (Empirical observation: the discarded layers actually include those with the highest compensation value, and removing their compensation leads to the most significant accuracy drop, to be detailed in Section~4.3).

\textbf{Truncated Pseudoinverse (Algorithm~1, lines 3--5).} We replace the naive inversion with a truncated pseudoinverse (truncated SVD). Setting a threshold $\tau = \text{rcond} \cdot \lambda_{\max}$, we apply truncation to the eigenvalues:
\begin{equation}
\Lambda^+_{ii} = 
\begin{cases} 
1/\lambda_i & \text{if } \lambda_i > \tau \\ 
0 & \text{if } \lambda_i \le \tau 
\end{cases}
\end{equation}
The compensation solution becomes:
\begin{equation}
 \begin{bmatrix} W_{\text{c}} \\ b_{\text{c}} \end{bmatrix} = V \Lambda^+ V^T \tilde{X}^T E 
\end{equation}
That is, we retain the directions above the threshold and take their reciprocals, while directly ignoring the directions below the threshold. The ignored directions contain no reliable information; thus, discarding them does not compromise effective compensation but eliminates numerical explosion, thereby recovering the high value layers erroneously discarded by existing methods. Note that the truncated pseudoinverse preserves the $R^2 \geq 0$ guarantee: the intercept direction of the augmented input has an eigenvalue that scales with the sample count and thus always survives truncation, so the constant predictor remains in the feasible set and $R^2 \geq 0$ holds by construction for our stable solution.

\textbf{Improvement stems from stability, not shrinkage.} We emphasize that this improvement originates from numerical stability rather than shrinkage after regularization. Unlike the biased Tikhonov/ridge regularization:
$$
\begin{aligned}
\text{ridge} &: (G + \alpha I)^{-1} \implies 1/\lambda_i \rightarrow \frac{1}{\lambda_i + \alpha} \ \ \text{for every direction} \\
\text{pinv}  &: \lambda_i \leq \tau \rightarrow 0, \quad \lambda_i > \tau \rightarrow 1/\lambda_i \ \ \text{(no shrinkage)}
\end{aligned}
$$
The truncated pseudoinverse (81.42\%) and ridge regularization (81.64\%) achieve almost identical accuracy, proving that the gain comes from discarding collapsed directions rather than from shrinkage, to be detailed in Section~4.2.

\subsection{Low Rank and Quantization Compression of the Compensator}

A full matrix compensator occupies a considerable model size, posing a burden for edge deployment. We compress the compensator along the axes of low rank decomposition and quantization (corresponding to Algorithm~1, lines 6--7).

\textbf{Axis 1: Activation aware low rank decomposition.} We approximate the full matrix compensator $W_{\text{c}}$ with low rank factors $A \cdot B$ (rank $r$). Unlike standard SVD truncation, which minimizes the parameter space error $\| W_{\text{c}} - W_{\text{r}} \|$, we choose to minimize the output space error:
$$ \min \| X(W_{\text{c}} - W_{\text{r}}) \|^2 \quad \text{subject to} \quad \text{rank}(W_{\text{r}}) \le r $$
By applying truncated SVD after whitening the Gram matrix, we obtain:
\begin{equation}
 W_{\text{r}} = G^{-1/2} \text{Tr}( G^{1/2} W_{\text{c}} ) 
\end{equation}
where $\text{Tr}$ denotes the SVD truncated to rank $r$, and $G^{1/2}$ is the symmetric square root of the Gram matrix. Here $G^{1/2}$ and $G^{-1/2}$ are computed from the truncated eigendecomposition of $G$, using the same rcond threshold as in the stable solver; directions below the threshold are excluded, so the whitening remains well defined even for the numerically singular layers identified in the previous section. Whitening concentrates the retained rank in the directions that activations actually pass through, minimizing the reconstruction error of the output. The resulting factors are $A$ (dimension $d_{\text{in}} \times r$) and $B$ (dimension $r \times d_{\text{out}}$), and during deployment, we compute $(x \cdot A) \cdot B$.

\textbf{Axis 2: Quantization of compensation factors.} We further apply channel wise symmetric quantization to the low rank factors $A$ and $B$ down to int8/int4. For each column $j$ of the factor ($b = \text{number of bits}$, $q_{\max} = 2^{b-1} - 1$):
\begin{equation}
\begin{aligned}
s &= \frac{\max |\text{column } j \text{ of } A|}{q_{\max}} \\
A_{\text{int}} &= \text{round}\left( \frac{A}{s} \right) \\
A &\approx s \cdot A_{\text{int}}
\end{aligned}
\end{equation}
We store the integer $A_{\text{int}}$ and the fp16 scale $s$. The bias $b_{\text{c}}$ is kept in full precision and is not quantized. The two axes are orthogonal and can be combined.

The aforementioned formulation of activation aware low rank compensation shares the same mathematical core as a series of recent training free low rank compensation methods (OLrC, QERA, CALDERA, EoRA, CLoQ). The contributions of this paper are: (i) integrating it into a comprehensive framework of layer wise and stable compensation; and (ii) systematically profiling the trade off between rank and bit width under the same size constraints.
\section{Experiments}

Our main experiments focus on ViT-B, and we verify the generalizability of the core mechanism on DeiT-T, Swin-T, and ResNet-50. All models are quantized under the W4A4 setting, using RepQ-ViT as the baseline quantizer. The compensation utilizes 512 training images as the calibration set, and the threshold for the truncated pseudoinverse is set to $10^{-3}$. All main results report the mean and standard deviation over 5 random seeds. Baseline methods for comparison include QwT, QwT* (which requires fine tuning), QwT-v2, and IGQ-ViT.

\subsection{Accuracy and Numerical Vulnerability}

\begin{figure}[!t]
\centering
\includegraphics[width=0.7\columnwidth]{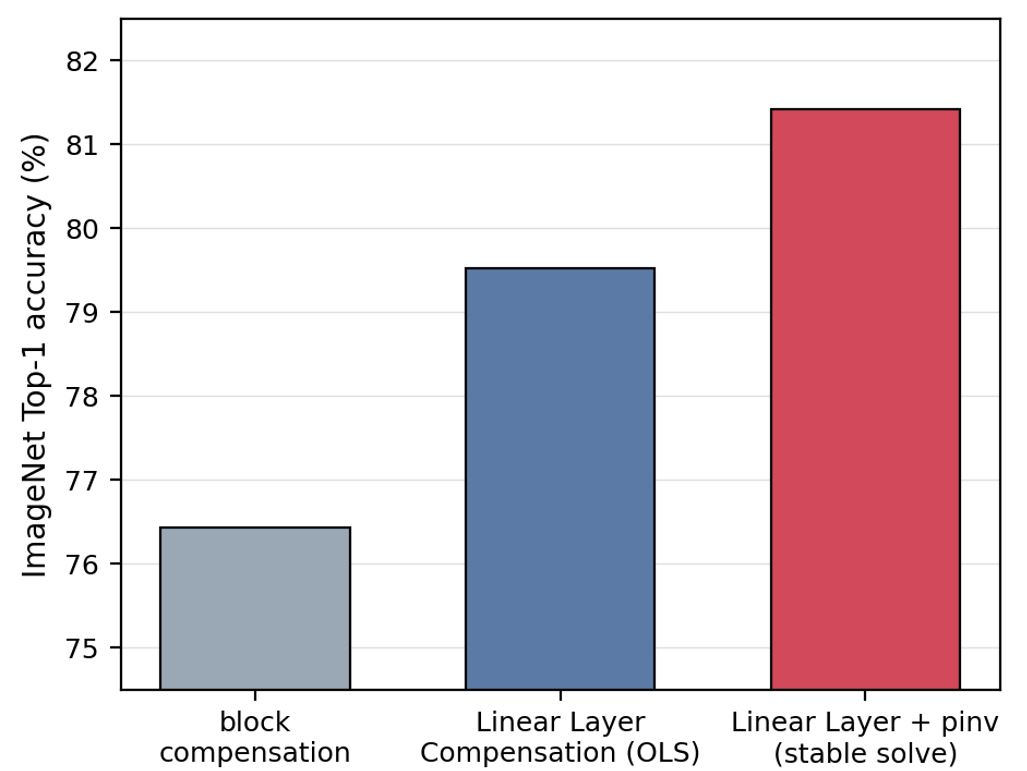}
\caption{Progressive accuracy improvements across the three stages of our method}
\label{fig2}
\end{figure}

Table \ref{main_results} presents the main results under ViT-B/W4A4. Our method achieves a Top 1 accuracy of 81.42\% $\pm$ 0.06 in a training free manner, outperforming the training free QwT (76.3\%), QwT* which requires additional fine tuning (78.5\%), QwT-v2 (75.6\%), and IGQ-ViT (79.3\%). The accuracy increases progressively across the three stages of our method: block wise compensation (76.43\%) $\rightarrow$ pushed down to the linear layer (79.52\%) $\rightarrow$ numerically stable solution (81.42\%) (Figure \ref{fig2} ). It is worth noting that the uncompressed full matrix compensator achieves 81.42\% accuracy but has a massive size of 215MB; after compression, it still maintains an accuracy of 80.26\% $\pm$ 0.11, while its size is reduced to 54.7MB.

\begin{table}[t]
\centering
\small
\begin{tabular}{l|c|c|c}
    \textbf{Method} & \textbf{Top 1 (\%)} & \textbf{Size (MB)} & \textbf{Training free} \\ \hline
    \makecell[l]{RepQ-ViT\\(baseline,\\uncompensated)} & 68.5 & 44.9 & --- \\ \hline
    QwT & 76.3 & 59.1 & \checkmark \\ \hline
    QwT* & 78.5 & 59.1 & $\times$ (Fine tuning) \\ \hline
    QwT-v2 & 75.6 & 45.6 & \checkmark \\ \hline
    IGQ-ViT & 79.3 & --- & --- \\ \hline
    \makecell[l]{Ours\\(Compressed)} & 80.26 $\pm$ 0.11 & 54.7 & \checkmark \\
\end{tabular}
\caption{Comparison of main results on ViT-B / W4A4 / ImageNet.}
\label{main_results}
\end{table}
\begin{table*}[t]
    \centering
    \footnotesize
    \renewcommand{\arraystretch}{0.8}
    \setlength{\tabcolsep}{4pt}
    \resizebox{0.96\textwidth}{!}{
    \begin{tabular}{l|l|l|l|l|l|p{4.5cm}}
        \textbf{Model} & \textbf{Arch.} & \textbf{Block level (QwT)} & \textbf{Direct Inverse} & \textbf{Ours (stable)} & \textbf{Neg-$R^2$ / Max Cond.} & \textbf{Failure Mode of Direct Inverse} \\ \hline
        ViT-B & ViT & 76.43 & 79.52$^\dagger$ (collapse 3/5) & pinv 81.42 $\pm$ 0.06 & 5 / $2.3\times 10^9$ & Divergent parameters; accuracy collapses to near random \\ \hline
        Swin-T & Swin & 74.72 & 78.40 $\pm$ 0.63 & pinv 78.90 $\pm$ 0.14 & 2--7 / $3.4\times 10^{10}$ & Unstable solution; excessively high variance \\ \hline
        DeiT-T & ViT & 61.40 & 64.66 $\pm$ 0.16 & pinv 64.76 $\pm$ 0.13 & 1--2 / $6.1\times 10^9$ & A few layers judged failed and skipped \\ \hline
        ResNet-50 & CNN & 62.5$^\ddagger$ & Fails (3/3) & ridge 68.72 / pinv 66.63 & singular / $\infty$ & Gram exactly singular; inversion undefined and fails \\
    \end{tabular}
    }
    \caption{Cross architecture numerical vulnerability. Ill conditioning is universal but manifests as distinct failure modes. $^\dagger$ViT-B direct inverse is reported at seed~0; averaging is uninformative as 3/5 seeds collapse to near random. Condition numbers indicate severity; collapse is driven by rank deficiency rather than large condition number alone (Appendix~A). $^\ddagger$ResNet-50 uses Percentile quantization (its QwT reference), not RepQ-ViT, so its baseline is not directly comparable to the transformer rows.}
    \label{cross_arch_vulnerability}
\end{table*}

The key to the improvement from 79.52\% to 81.42\% lies in resolving the numerical collapse during the compensation solving process. Table \ref{solving_methods} compares three solving methods: direct inversion collapses in 3 out of 5 seeds, with the accuracy of the collapsed seeds dropping to near random levels (approximately 0.1\%). Its 79.52\% is merely a lucky outcome from the uncollapsed seeds, whereas the truncated pseudo inverse remains stable across all 5 seeds, achieving 81.42\% $\pm$ 0.06. We further verify that this improvement stems from numerical stability rather than the shrinkage of regularization. Therefore, we also test the ridge method, which achieves an almost identical 81.64\% $\pm$ 0.04, indicating that the gain comes from discarding the collapsed directions rather than the universal shrinkage applied by ridge across all directions.

\begin{table}[t]
\centering
\footnotesize
\renewcommand{\arraystretch}{0.8}
\setlength{\tabcolsep}{4pt}
\begin{tabular}{l|c|c|c}
    \makecell{\textbf{Method}} &
    \makecell{\textbf{Top 1}\\\textbf{(\%)}} &
    \makecell{\textbf{Collapsed}\\\textbf{Seeds}} &
    \makecell{\textbf{Layers with}\\\textbf{Negative $R^2$}} \\ \hline
    \makecell[l]{OLS\\(direct inverse)} &
    79.52 &
    3 / 5 &
    5 \\ \hline
    \makecell[l]{Truncated\\pseudo inverse} &
    81.42 $\pm$ 0.06 &
    0 / 5 &
    0 \\ \hline
    \makecell[l]{Ridge\\(Tikhonov)} &
    81.64 $\pm$ 0.04 &
    0 / 5 &
    0 \\
\end{tabular}
\caption{Comparison of solving methods for compensation.}
\label{solving_methods}
\end{table}

\begin{figure*}[!t]
    \centering
    \includegraphics[width=0.7\textwidth]{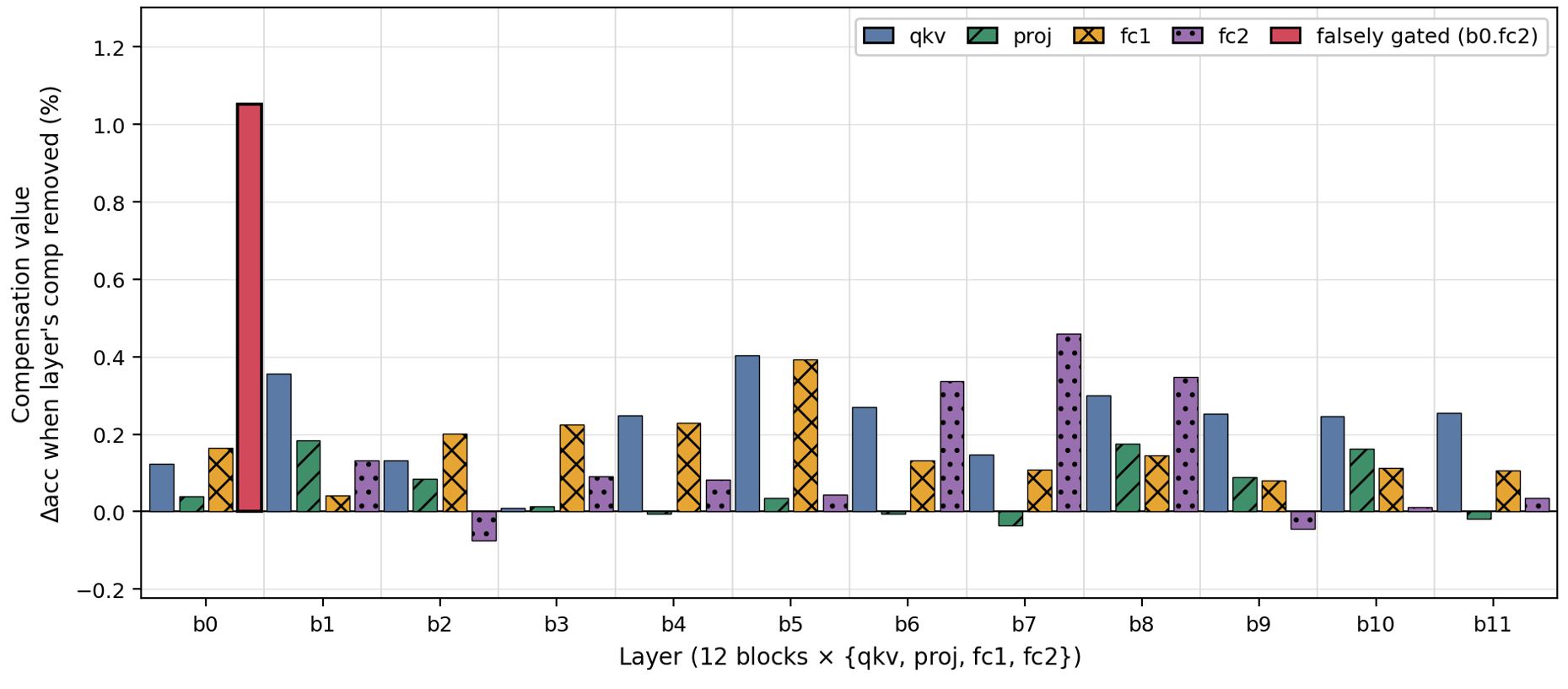}
    \caption{Compensation Value Plot}
    \label{fig3}
\end{figure*}

\begin{figure*}[!t]
    \centering
    \includegraphics[width=0.7\textwidth]{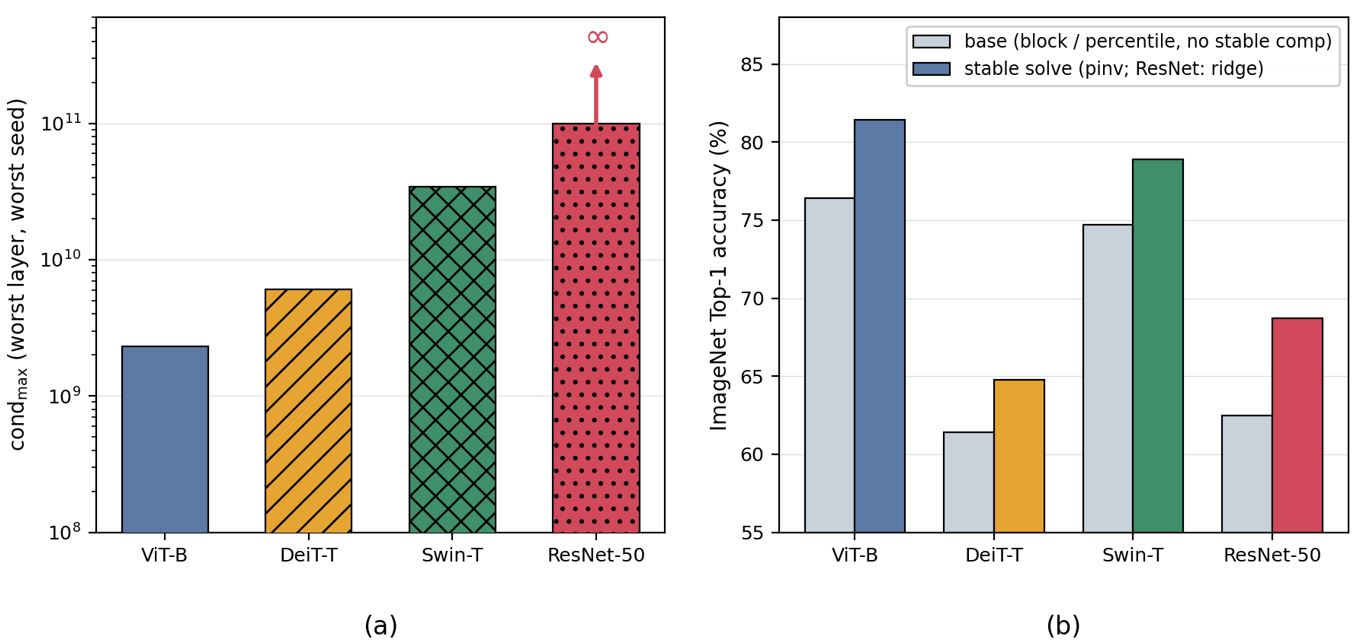}
    \caption{(a)Severity of ill conditioning across architectures (b)Accuracy comparison before and after stabilization.}
    \label{fig4}
\end{figure*}

\begin{table}[t]
    \centering
    \footnotesize
    \renewcommand{\arraystretch}{0.8}
    \setlength{\tabcolsep}{4pt}
    \begin{tabular}{l|c|c|c}
        \textbf{Rank / Bits} & \textbf{16-bit} & \textbf{8-bit} & \textbf{4-bit} \\ \hline
        $r=256$ & 81.14 / 119.0 & 81.16 / 82.3 & 80.79 / 63.9 \\ \hline
        $r=128$ & 80.79 / 82.4 & 80.67 / 64.0 & 80.50 / 54.7 \\ \hline
        $r=64$ & 80.34 / 64.0 & 80.37 / 54.8 & 79.87 / 50.1 \\
    \end{tabular}
    \caption{Relationship between rank and quantization bits. Performance is reported as Top-1 Accuracy (\%) / Size (MB).}
    \label{rank_and_bits}
\end{table}

\begin{figure}[!t]
\centering
\includegraphics[width=0.7\columnwidth]{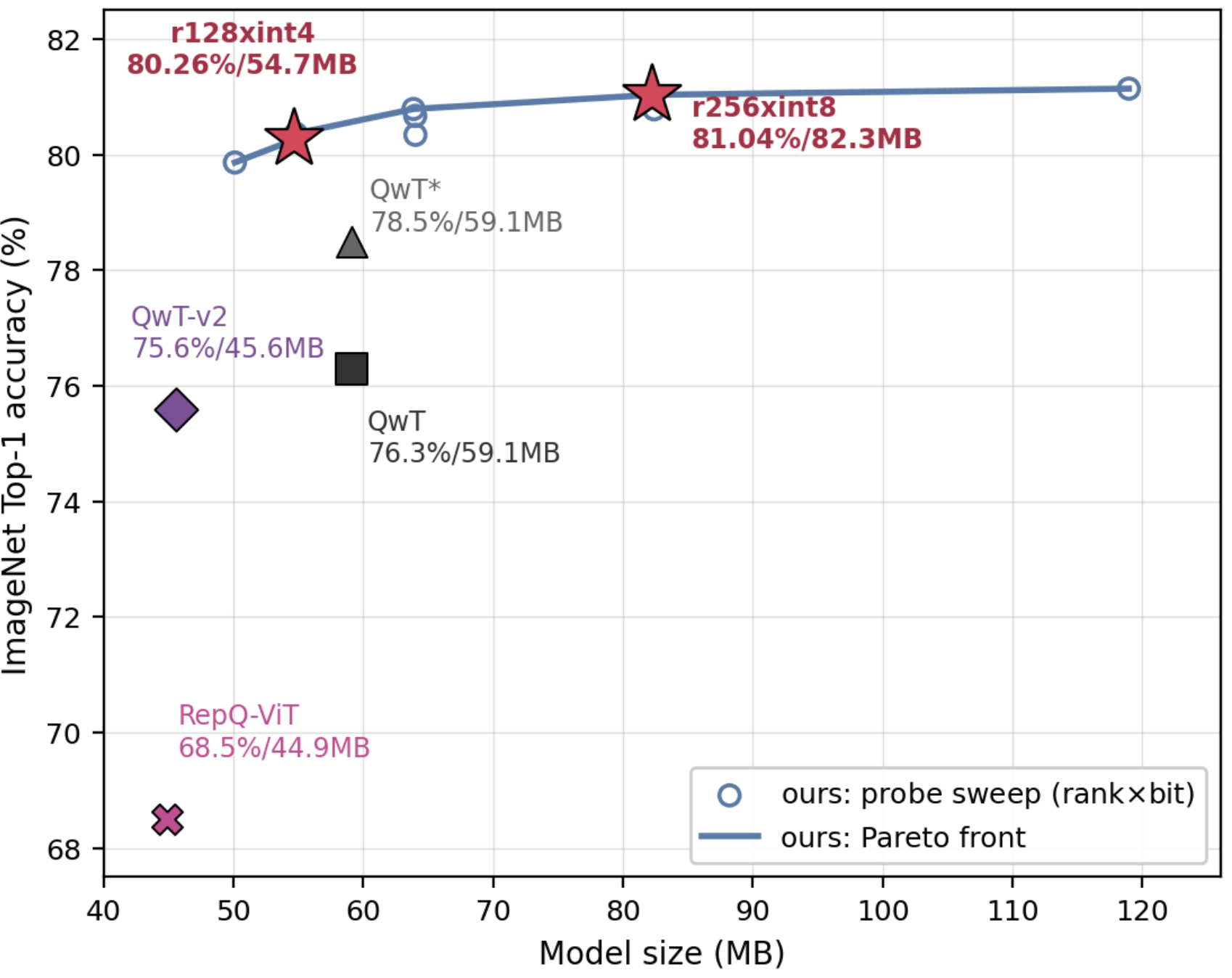}
\caption{Trade off plot between accuracy and size}
\label{fig5}
\end{figure}

Under existing methods, the direct solving approach is misjudged as uncompensable and discarded by the $R^2 > 0$ threshold. We quantify the contribution of each layer's compensation through a compensation value ablation study (Figure \ref{fig3}). The results reveal that b0.fc2 is the most valuable among all 48 compensators; removing its compensation results in a 1.05\% accuracy drop, significantly higher than that of all other layers. Ironically, it is precisely this most valuable layer that is misjudged as ``uncompensable'' and discarded by the $R^2 > 0$ threshold---and this is exactly the layer we salvage through our stable solution.

\subsection{Cross Architecture Analysis}

All four architectures exhibit ill conditioned Gram matrices during
the compensation solving process, albeit to varying degrees of
severity: Figure \ref{fig4}(a) shows that the maximum condition number
increases progressively across architectures, from $2.3 \times 10^9$
for ViT-B, to DeiT-T, Swin-T, and finally ResNet-50. The failure
manifests differently: silent collapse of the least squares solution
(ViT-B), high variance and the performance degradation of individual
seeds (Swin-T), 1--2 layers discarded by the $R^2 > 0$ threshold
(DeiT-T), and the direct inversion throwing an exception and crashing
outright (ResNet-50). Yet numerical stabilization proves effective
across all four architectures: Figure \ref{fig4}(b) demonstrates that the
stable solution brings consistent accuracy gains over the unstable
baseline. It must be emphasized that what is verified here is the
universality of the core mechanism of numerical stabilization, rather
than claiming that the complete compensation framework achieves the
optimal solution on any arbitrary architecture.

We do not claim that any single form of stabilization is universally optimal; rather, we argue that numerical stabilization itself is necessary, and this numerical vulnerability has not yet been fully addressed in existing compensation implementations.

\subsection{Uniform Allocation is Sufficient}

A natural idea is to allocate compensation resources in a fine grained manner based on the importance of each position. We verify through ablation studies that such position wise allocation is not superior to uniform allocation. Table 10 in Appendix compares different rank allocation strategies under the same size constraint (54.7MB). Uniform allocation achieves 80.376\%, whereas allocating higher ranks to important layers, allocating lower ranks to important layers, or adjusting continuously based on importance all yield slightly lower accuracy than uniform allocation.

\subsection{Compression Results}
Figure~\ref{fig5} presents the accuracy--size trade off. Our primary
configuration (80.26\% $\pm$ 0.11 / 54.7MB) dominates QwT, QwT-v2, and
the fine tuning required QwT$^*$ in both dimensions while remaining
training free, leading QwT$^*$ by 1.76\% at 4.4MB smaller size.
Table~\ref{rank_and_bits} shows that under a fixed size constraint, a
higher rank paired with more aggressive quantization outperforms a lower
rank paired with milder quantization, indicating that the compensation
benefits more from retained subspace dimensionality than from per parameter precision.

\section{Conclusion and Future Work}
This paper reveals a problem overlooked by existing training free residual
compensation methods: activation quantization drives the Gram matrix on
which the compensation solver relies past the threshold of numerical
collapse, and the collapsed layers are then falsely gated as
``uncompensable'' by the standard $R^2 > 0$ criterion. This gate
ironically discards the layers with the highest compensation value. We
replace direct inversion with a truncated pseudoinverse to salvage them,
verify that the gain originates from numerical stability rather than
regularization induced shrinkage, and show that this vulnerability is not
an isolated case. Built on this stable solver, our layer wise framework
reaches 81.42\% Top-1 on ViT-B / W4A4 / ImageNet without training, and
compresses to a deployable operating point of 80.26\% at 54.7MB that
outperforms both QwT and QwT-v2 as well as the fine tuning required
QwT$^*$ in accuracy and size.

Three directions remain open. First, while numerical ill conditioning
spans both CNNs and Transformers and the stable solution rescues the
solver on ResNet-50, applying the complete layer wise framework to CNNs
remains an open challenge: module level compensation does not yet stably
outperform the block level baseline, which we attribute to the error
propagation structure of convolutional layers. Second, ill conditioning
worsens as bit width decreases, but at lower bit widths such as W3A3 the
baseline itself degrades severely, so stabilizing both simultaneously
remains unresolved. Third, the proposed solver is in principle agnostic
to the specific quantizer, and combining it with other W4A4 quantizers is
a natural extension.

\bibliographystyle{unsrtnat}
\bibliography{references}

@inproceedings{liu2024llm,
  title={Llm-qat: Data-free quantization aware training for large language models},
  author={Liu, Zechun and Oguz, Barlas and Zhao, Changsheng and Chang, Ernie and Stock, Pierre and Mehdad, Yashar and Shi, Yangyang and Krishnamoorthi, Raghuraman and Chandra, Vikas},
  booktitle={Findings of the Association for Computational Linguistics: ACL 2024},
  pages={467--484},
  year={2024}
}

@inproceedings{chen2025efficientqat,
  title={Efficientqat: Efficient quantization-aware training for large language models},
  author={Chen, Mengzhao and Shao, Wenqi and Xu, Peng and Wang, Jiahao and Gao, Peng and Zhang, Kaipeng and Luo, Ping},
  booktitle={Proceedings of the 63rd Annual Meeting of the Association for Computational Linguistics (Volume 1: Long Papers)},
  pages={10081--10100},
  year={2025}
}

@inproceedings{zhao2023post,
  title={Post-training quantization or quantization-aware training? That is the question},
  author={Zhao, Xiaotian and Xu, Ruge and Guo, Xinfei},
  booktitle={2023 China Semiconductor Technology International Conference (CSTIC)},
  pages={1--3},
  year={2023},
  organization={IEEE}
}

@article{li2021brecq,
  title={Brecq: Pushing the limit of post-training quantization by block reconstruction},
  author={Li, Yuhang and Gong, Ruihao and Tan, Xu and Yang, Yang and Hu, Peng and Zhang, Qi and Yu, Fengwei and Wang, Wei and Gu, Shi},
  journal={arXiv preprint arXiv:2102.05426},
  year={2021}
}

@article{arai2026quantization,
  title={Quantization error propagation: Revisiting layer-wise post-training quantization},
  author={Arai, Yamato and Ichikawa, Yuma},
  journal={Advances in Neural Information Processing Systems},
  volume={38},
  pages={151916--151951},
  year={2026}
}

@article{bai2022towards,
  title={Towards efficient post-training quantization of pre-trained language models},
  author={Bai, Haoli and Hou, Lu and Shang, Lifeng and Jiang, Xin and King, Irwin and Lyu, Michael R},
  journal={Advances in neural information processing systems},
  volume={35},
  pages={1405--1418},
  year={2022}
}

@inproceedings{zhang2026quantvla,
  title={Quantvla: Scale-calibrated post-training quantization for vision-language-action models},
  author={Zhang, Jingxuan and Hsieh, Yunta and Wan, Zhongwei and Lin, Haokun and Wang, Xin and Wang, Ziqi and Lei, Yingtie and Zhang, Mi},
  booktitle={Proceedings of the IEEE/CVF Conference on Computer Vision and Pattern Recognition},
  pages={39539--39549},
  year={2026}
}

@inproceedings{shang2023post,
  title={Post-training quantization on diffusion models},
  author={Shang, Yuzhang and Yuan, Zhihang and Xie, Bin and Wu, Bingzhe and Yan, Yan},
  booktitle={Proceedings of the IEEE/CVF conference on computer vision and pattern recognition},
  pages={1972--1981},
  year={2023}
}

@article{tang2025qwt,
  title={Qwt-v2: Practical, effective and efficient post-training quantization},
  author={Tang, Ningyuan and Fu, Minghao and Yu, Hao and Wu, Jianxin},
  journal={arXiv preprint arXiv:2505.20932},
  year={2025}
}

@article{frantar2022gptq,
  title={Gptq: Accurate post-training quantization for generative pre-trained transformers},
  author={Frantar, Elias and Ashkboos, Saleh and Hoefler, Torsten and Alistarh, Dan},
  journal={arXiv preprint arXiv:2210.17323},
  year={2022}
}

@article{hastie2020ridge,
  title={Ridge regularization: An essential concept in data science},
  author={Hastie, Trevor},
  journal={Technometrics},
  volume={62},
  number={4},
  pages={426--433},
  year={2020},
  publisher={Taylor \& Francis}
}

@inproceedings{zhang2025qera,
  title={Qera: an analytical framework for quantization error reconstruction},
  author={Zhang, Cheng and Wong, Jeffrey TH and Xiao, Can and Constantinides, George and Zhao, Yiren},
  booktitle={International Conference on Learning Representations},
  volume={2025},
  pages={12531--12560},
  year={2025}
}

@inproceedings{fu2025quantization,
  title={Quantization without tears},
  author={Fu, Minghao and Yu, Hao and Shao, Jie and Zhou, Junjie and Zhu, Ke and Wu, Jianxin},
  booktitle={Proceedings of the Computer Vision and Pattern Recognition Conference},
  pages={4462--4472},
  year={2025}
}

@inproceedings{mohammadi2026fixing,
  title={Fixing Quantization with Lightweight Adapters},
  author={Mohammadi, Mohammadreza and Grenier, Matthew and Zand, Ramtin},
  booktitle={Proceedings of the IEEE/CVF Conference on Computer Vision and Pattern Recognition},
  pages={3569--3578},
  year={2026}
}

@inproceedings{moon2024instance,
  title={Instance-aware group quantization for vision transformers},
  author={Moon, Jaehyeon and Kim, Dohyung and Cheon, Junyong and Ham, Bumsub},
  booktitle={Proceedings of the IEEE/CVF conference on computer vision and pattern recognition},
  pages={16132--16141},
  year={2024}
}

@inproceedings{li2023repq,
  title={Repq-vit: Scale reparameterization for post-training quantization of vision transformers},
  author={Li, Zhikai and Xiao, Junrui and Yang, Lianwei and Gu, Qingyi},
  booktitle={Proceedings of the IEEE/CVF International Conference on Computer Vision},
  pages={17227--17236},
  year={2023}
}

@inproceedings{wu2025fima,
  title={FIMA-Q: Post-training quantization for vision transformers by fisher information matrix approximation},
  author={Wu, Zhuguanyu and Wang, Shihe and Zhang, Jiayi and Chen, Jiaxin and Wang, Yunhong},
  booktitle={Proceedings of the Computer Vision and Pattern Recognition Conference},
  pages={14891--14900},
  year={2025}
}

@inproceedings{wu2025aphq,
  title={Aphq-vit: Post-training quantization with average perturbation hessian based reconstruction for vision transformers},
  author={Wu, Zhuguanyu and Zhang, Jiayi and Chen, Jiaxin and Guo, Jinyang and Huang, Di and Wang, Yunhong},
  booktitle={Proceedings of the Computer Vision and Pattern Recognition Conference},
  pages={9686--9695},
  year={2025}
}

@inproceedings{li2024loftq,
  title={Loftq: Lora-fine-tuning-aware quantization for large language models},
  author={Li, Yixiao and Yu, Yifan and Liang, Chen and Karampatziakis, Nikos and He, Pengcheng and Chen, Weizhu and Zhao, Tuo},
  booktitle={International Conference on Learning Representations},
  volume={2024},
  pages={13409--13424},
  year={2024}
}

@article{saha2024compressing,
  title={Compressing large language models using low rank and low precision decomposition},
  author={Saha, Rajarshi and Sagan, Naomi and Srivastava, Varun and Goldsmith, Andrea J and Pilanci, Mert},
  journal={Advances in Neural Information Processing Systems},
  volume={37},
  pages={88981--89018},
  year={2024}
}

@article{zhang2026q,
  title={Q-DiT4SR: Exploration of Detail-Preserving Diffusion Transformer Quantization for Real-World Image Super-Resolution},
  author={Zhang, Xun and Yang, Kaicheng and Lu, Hongliang and Qin, Haotong and Guo, Yong and Zhang, Yulun},
  journal={arXiv preprint arXiv:2602.01273},
  year={2026}
}

\end{document}